\documentclass[11pt]{article}
\usepackage[final]{acl}
\usepackage{times}
\usepackage{latexsym}
\usepackage[T1]{fontenc}
\usepackage[utf8]{inputenc}
\usepackage{inconsolata}
\usepackage{graphicx}
\usepackage{float}
\usepackage{url}
\usepackage[table]{xcolor}
\usepackage{booktabs}
\usepackage{multirow}
\usepackage{array}
\usepackage{siunitx}
\usepackage{caption}
\usepackage{arydshln}
\usepackage{nicematrix}
\usepackage{tcolorbox}
\tcbuselibrary{breakable}
\usepackage{setspace}
\usepackage{xcolor}
\usepackage{bbding}
\usepackage{algorithm}
\usepackage{algpseudocode}
\usepackage{graphicx}
\usepackage{booktabs}
\usepackage{wrapfig}
\usepackage{subcaption}
\usepackage{adjustbox}
\usepackage{enumitem}
\usepackage{amsmath}
\usepackage{amssymb}
\usepackage{hyperref}
\usepackage{cleveref}
\usepackage[nopatch=footnote]{microtype}
\definecolor{light_purple}{RGB}{235,236,242}
\definecolor{light_blue}{RGB}{244,249,254}
\definecolor{bgcolor}{RGB}{242, 243, 245}

\newcolumntype{B}{>{\columncolor{blue!4}}c}
\newcolumntype{R}{>{\columncolor{red!4}}c}
\newcolumntype{d}{>{\columncolor{brown!4}}c}
\newcolumntype{q}{>{\columncolor{green!4}}c}
\newcolumntype{P}{>{\columncolor{purple!4}}c}
\newcolumntype{Y}{>{\columncolor{yellow!4}}c}

\title{Don't Tell the Answer, Truly Guide the Reasoning During RL Rollouts}

\author{
    \textbf{Xinyi Wang\textsuperscript{\rm 1}, Jinyi Han\textsuperscript{\rm 2}, Zishang Jiang\textsuperscript{\rm 1}, Tingyun Li\textsuperscript{\rm 1}, Jiaqing Liang\textsuperscript{\rm 1}\thanks{Corresponding author.},}\\
    \textbf{Sihang Jiang\textsuperscript{\rm 3}, Zhaoqian Dai\textsuperscript{\rm 4}, Ma Shuguang\textsuperscript{\rm 4}, Fei Yu\textsuperscript{\rm 4}, Yanghua Xiao\textsuperscript{\rm 3}} \\
    \textsuperscript{\rm 1}School of Data Science, Fudan University\\
    \textsuperscript{\rm 2}Shanghai Institute of Artificial Intelligence for Education, East China Normal University\\
    \textsuperscript{\rm 3}College of Computer Science and Artificial Intelligence, Fudan University \\
    \textsuperscript{\rm 4}Ant Group \\
    \texttt{xinywang24@m.fudan.edu.cn}
}

\begin{document}

\maketitle
\begin{abstract}
Reinforcement Learning (RL) has become a key driver for enhancing the long chain-of-thought (CoT) reasoning capabilities of Large Language Models (LLMs). 
However, prevalent methods like GRPO often fail when task difficulty exceeds model capacity, leading to reward sparsity and inefficient training. 
Prior work attempts to mitigate this with off-policy data, but such methods often induce severe distributional mismatches that destabilize policy updates.
In this work, we identify a core issue underlying these failures, which we term low training affinity, and introduce \textit{Affinity}, the first quantitative metric for monitoring the compatibility between external guidance and the model's intrinsic policy.
To address this, we propose HINT, an adaptive framework designed to enhance reasoning capabilities while explicitly preserving high \textit{Affinity}.
First, instead of revealing partial answers, HINT supplies \textbf{Meta-Hints}, which act as abstract cognitive scaffolding to guide the model in articulating solutions independently.
Second, to ensure stability, we integrate \textbf{Affinity-Aware Policy Optimization (AAPO)}, which dynamically modulates the learning objective based on the \textit{Affinity}.
Extensive experiments across diverse benchmarks demonstrate that HINT consistently outperforms strong baselines, while exhibiting superior stability and robust generalization to out-of-distribution tasks.
Code are available at Github\footnote{\url{https://github.com/ViviqwerAsd/HINT}}.
\end{abstract}

\section{Introduction}

\begin{figure}[ht]
  \centering
  \includegraphics[width=\columnwidth]{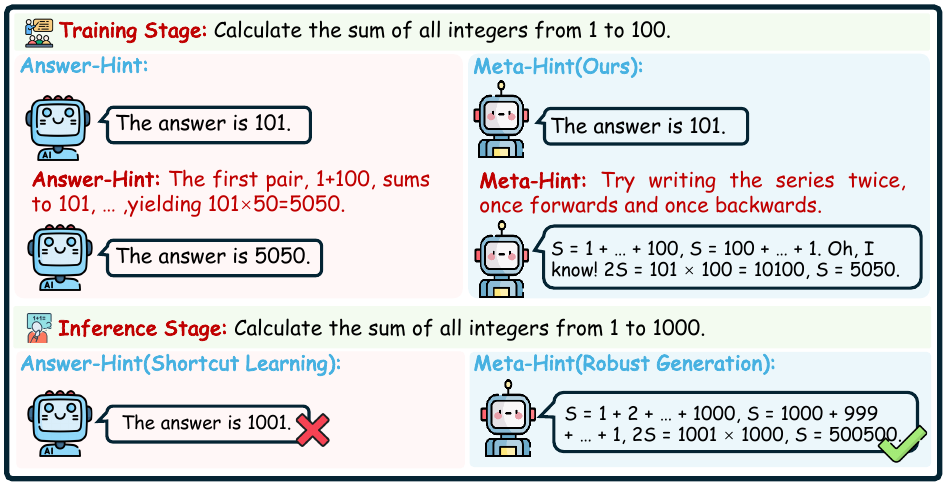}
  \caption{Comparison of Hint Mechanisms and Their Impact on Learning. \textbf{Left:} Answer-Hints provide explicit partial solutions. The model maximizes rewards by simply completing this pre-defined path, which leads to \textbf{Shortcut Learning}, characterized by the memorization of surface patterns rather than an understanding of the underlying logic. \textbf{Right:} In contrast, our Meta-Hints offer high-level cognitive scaffolding, \textbf{compelling the model to develop solution path independently} and fostering robust generation.}
  \label{fig:intro_hint}
\end{figure}

RL methods, particularly GRPO~\citep{shao2024deepseekmath}, play a pivotal role in advancing long CoT reasoning~\citep{wei2022chain}.
By avoiding the instability and overhead of training a separate value model, GRPO leverages group-based reward aggregation to deliver stable and efficient learning signals.
Such RL approaches~\citep{ahmadian2024back, shao2024deepseekmath, hu2025reinforce++, yu2025dapo} have become a key driver of progress in reasoning ability, enabling models to explore solution paths on verifiable problems.
Building on these advances, recent reasoning models such as DeepSeek-R1~\citep{guo2025deepseek} and OpenAI-o1~\citep{jaech2024openai} have achieved remarkable performance on complex tasks like mathematical reasoning and code generation~\citep{jiang2024survey}.

A critical challenge for GRPO, despite its strong empirical performance, is its tendency to generate sample groups consisting entirely of incorrect answers on tasks whose difficulty exceeds policy-model capacity, resulting in reward sparsity~\citep{zhao2025echo, yue2025does}.
This sparsity renders gradients uninformative and leaves the policy model without effective learning signal, reducing training efficiency and wasting valuable data.

\textbf{Leveraging external, off-policy data is a key method for addressing this issue.}
This method has been implemented in prior work through two main lines of remedies.
(I) \textbf{Mixed-policy}~\citep{yan2025learning,zhang2025policy,fu2025srft}: Mixed-policy interleaves RL with externally provided high-quality reference trajectories to stabilize training.
(II) \textbf{Using hints}~\citep{li2025questa, liu2025ghpo, zhang2025bread}: To mitigate reward sparsity and ensure continuous training updates, another common approach is to expose the model to partial solution content during rollout, guiding exploration along correct trajectories.

Despite their potential benefits, both approaches share a common limitation: they inject complete or partial solution content derived from reference trajectories that can deviate substantially from the model's current policy.
We collectively refer to such external solution content as \textit{Answer-Hints}.
We refer to the compatibility between these Answer-Hints and the model's intrinsic policy as \textit{training affinity}.
When this affinity is low, importance sampling ratios become highly variable, leading to unstable gradients and deceptive learning signals~\citep{yan2025learning}, as illustrated in Figure~\ref{fig:rewards_accuracy}.
To make this notion measurable, we draw on PPO's clipping mechanism as a proxy for update stability~\citep{schulman2017proximal} and introduce \textit{Affinity}, a metric that quantifies training affinity in terms of clipping frequency and update consistency.

To leverage off-policy data while preserving high \textit{Affinity}, the key is to \textbf{guide the model toward discovering the solution on its own, rather than revealing the solution path explicitly.}
To this end, we propose HINT, an adaptive framework for leveraging off-policy data while preserving high \textit{Affinity}. HINT does so by providing heuristic \textit{Meta-Hints}, namely high-level scaffolding that steers exploration without disclosing partial answers.
Akin to the Socratic method, such guidance encourages the model to navigate challenges independently and thereby develop more robust reasoning behaviors.
To ensure that these guided updates translate into stable policy improvement, we further introduce Affinity-Aware Policy Optimization (AAPO), which dynamically modulates the objective according to the compatibility between the guidance and the model's intrinsic distribution.

Our contributions are summarized as follows:
\begin{itemize}[nosep]
    \item We formally define low training affinity as a key failure mode when integrating off-policy data into on-policy RL frameworks, and propose \textit{Affinity}, a quantitative metric to monitor these dynamics.
    \item We propose the HINT framework, which synergizes Meta-Hints to guide the model in discovering effective reasoning paths independently, and AAPO to ensure training stability.
    \item Extensive experiments demonstrate that HINT consistently outperforms Answer-Hints baselines across in-domain and out-of-domain benchmarks, exhibiting superior robustness.
\end{itemize}

\section{Related Work}
\subsection{Reinforcement Learning for Large Language Model Reasoning.}
Recent advances in RL approaches have significantly enhanced the reasoning capabilities of LLMs. 
Large reasoning Models (LRMs) such as OpenAI-o1~\citep{jaech2024openai}, DeepSeek-R1~\citep{guo2025deepseek}, and Kimi-1.5~\citep{team2025kimi} achieve state-of-the-art performance on complex reasoning tasks (e.g., mathematics, coding, scientific problem solving) by leveraging Reinforcement Learning from Verifiable Rewards (RLVR)~\citep{liu2025understanding, hu2025open, cui2025process}, where automatically checkable rules provide supervision signals. 
Compared to earlier methods like SFT or reinforcement learning from human feedback (RLHF), RLVR has shown superior generalization and robustness~\citep{chu2025sft, snell2025scaling}. 
Building on this paradigm, subsequent studies have proposed improved optimization strategies and structured prompting techniques that further strengthen reasoning capabilities~\citep{schulman2017proximal,wang2020truly}. 
Despite this progress, a critical failure mode for existing RL methods is reward sparsity, which occurs when all rollouts in a sample fail.
Overcoming this challenge is essential for enhancing the stability and sample efficiency of training.

\subsection{Improving Rollout Efficiency in RL for LLMs.} 
A well-known challenge in methods such as GRPO is the vanishing gradient issue. 
This problem occurs when all trajectories in a sample group are incorrect, as the group advantage collapses to zero, yielding no gradient for policy updates~\citep{shao2024deepseekmath,guo2025deepseek}. 
To mitigate this, some works have focused on injecting external, off-policy data to improve training efficiency and stability. 
This has been explored through two main strategies. 
Some methods use mixed-policy, replacing a portion of on-policy rollouts with complete, high-quality reference trajectories from off-policy datasets~\citep{yan2025learning,lin2025cppo,xu2025not,wang2025dump}. 
Others employ partial supervision, providing segments of a reference solution to rescue failed rollouts~\citep{li2025questa, liu2025ghpo, zhang2025bread}.
In this work, we collectively view these forms of complete or partial reference-solution exposure as \textit{Answer-Hints}.
While these approaches effectively improve rollout efficiency, their over-reliance on off-policy data can misguide policy updates, steering the model toward non-generalizable or spurious solution paths.

\section{Methods}
\label{sec:hint}

\subsection{Preliminary}
\label{sec:preliminary}
Following recent work~\citep{yu2025dapo, yan2025learning}, we build upon GRPO~\citep{guo2025deepseek} and omit the KL penalty term.
For each prompt, GRPO draws a group of $n$ rollouts and computes a group-normalized advantage for every token.
Mathematically, GRPO optimizes the behavior of model through the following objective function:
\begin{equation}
\small
\label{grpo}
\begin{aligned}
\mathcal{J}_{\text{GRPO}}(\theta) &= 
\mathbb{E}_{q \sim Q, \{y^{(i)}\}_{i=1}^n \sim \pi_{\text{old}}(\cdot|q)}
\frac{1}{n} \sum_{i=1}^n 
\frac{1}{|y^{(i)}|} \sum_{t=1}^{|y^{(i)}|}  \\
&\hspace{-1em}
\min \left[
r_{t}^{(i)}(\theta) \hat{A}_{t}^{(i)}, 
\text{clip}(r_{t}^{(i)}(\theta), 1 \pm \epsilon) \hat{A}_{t}^{(i)}
\right]  
,
\end{aligned}
\end{equation}
where \( r_{t}^{(i)}(\theta) = \frac{\pi_\theta(y_{t}^{(i)} \mid q, y_{<t}^{(i)})}{\pi_{\text{old}}(y_{t}^{(i)} \mid q, y_{<t}^{(i)})} \) 
is the importance sampling ratio between the current policy and the behavior policy.
 
Let $\{R_i\}_{i=1}^n$ denote the sequence-level rewards assigned to these rollouts. 
The token-level advantages $\hat{A}_{t}^{(i)}$ are computed by normalizing each trajectory's reward within the group:
\begin{equation}
\hat{A}_{t}^{(i)}
=
\frac{R_i - \mathrm{mean}(\{R_j\}_{j=1}^n)}
{\mathrm{std}(\{R_j\}_{j=1}^n) + \varepsilon}.
\label{eq:grpo_advantage}
\nonumber
\end{equation}

When all rollouts in a group are assigned identical rewards, 
$R_i - \mathrm{mean}(\{R_j\}_{j=1}^n)$ becomes zero for every $i$, causing every advantage $\hat{A}_{t}^{(i)}$ to collapse to zero. 
Such prompts therefore provide no learning signal during training.
Conversely, prompts that produce non-identical rewards across the group yield non-zero advantages and therefore generate meaningful gradients.

\subsection{Quantifying the Quality of Exploration}
\label{sec:metrics}
\begin{figure}[h]
  \centering
  \includegraphics[width=\linewidth]{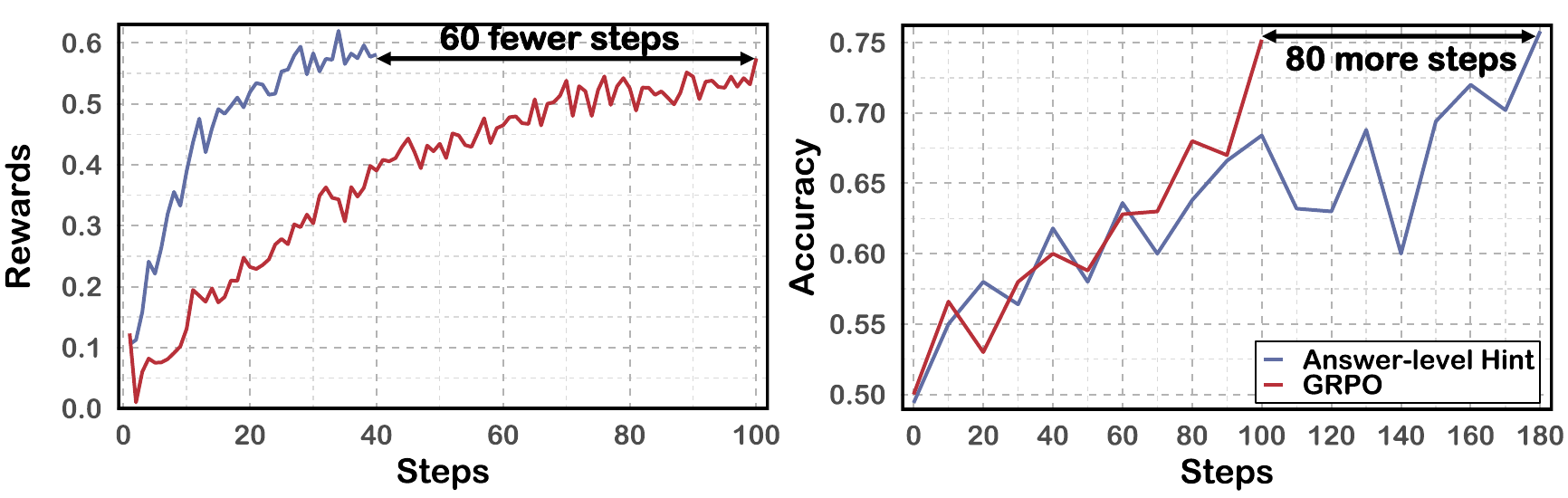}
  \caption{The \textit{Illusion} of High Rewards During Training with the Answer-Hints Method.
    \textbf{Left:} Training rewards surge rapidly. 
    \textbf{Right:} Test accuracy on MATH-500 stagnates. 
    This discrepancy indicates that reward signals alone cannot reliably represent the actual training state.}
  \label{fig:rewards_accuracy}
\end{figure}
While strategies like Answer-Hints mitigate sparsity, they often induce the ``Illusion of High Rewards'', a phenomenon where training rewards surge while generalization stagnates (Figure~\ref{fig:rewards_accuracy}).
This discrepancy arises because strong external guidance creates distributional mismatches, inflating reward metrics while yielding uninformative or unstable gradients.
Consequently, relying solely on rewards is deceptive. To capture the true training dynamics, we must look beyond reward accumulation and introduce rigorous metrics that quantify both the effectiveness and stability of policy updates.

\begin{figure*}[ht]
  \centering
  \includegraphics[width=0.95\textwidth]{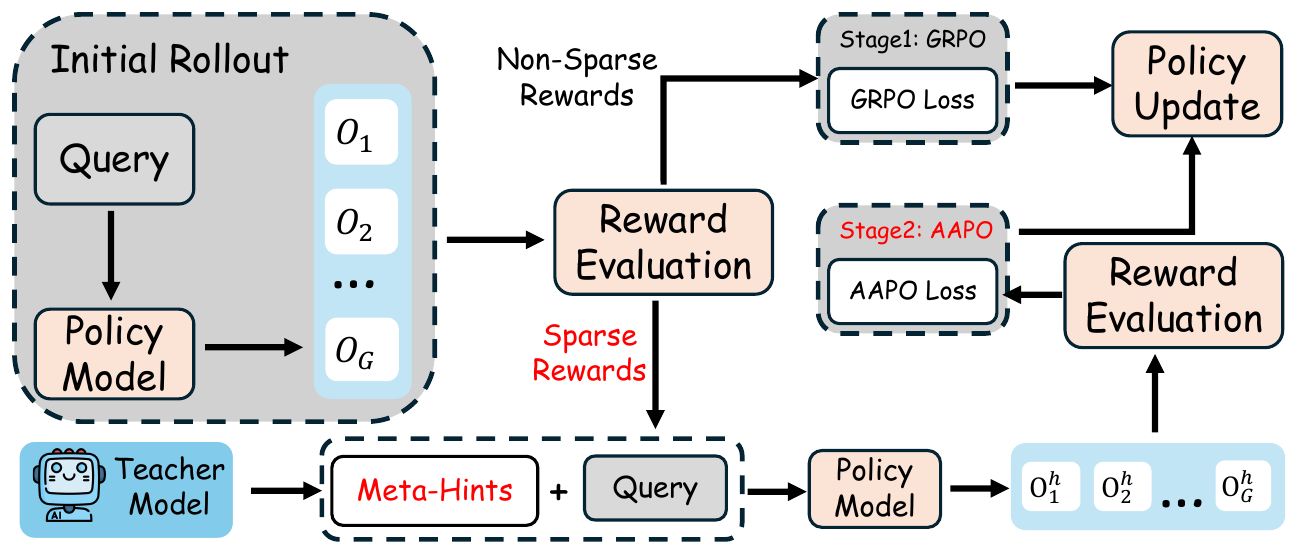}
  \caption{The HINT Framework: An Adaptive Two-Stage Rollout Process. HINT operates in two stages. 
\textbf{(I) Standard Rollout:} The model first samples trajectories from the original problem. If the rewards are non-sparse, the process follows the standard GRPO update path. 
\textbf{(II) Hint-Augmented Rescue:} If rewards are sparse (all trajectories are incorrect), the HINT mechanism is activated. The model re-rolls out conditioned on a \textbf{Meta-Hint} to guide exploration toward correct solutions. Crucially, to mitigate the potential instability introduced by external guidance, these updates are optimized via \textbf{Affinity-Aware Policy Optimization (AAPO)}, which dynamically gates gradients based on update affinity to filter out noise.}
  \label{fig:framework}
\end{figure*}

\textbf{Effective Update Ratio (EUR).}
\textbf{EUR quantifies how many token-level updates remain unclipped under the clipped objective.}
Recall from Eq.~\eqref{grpo} that GRPO generates token-level advantages $\hat{A}_{t}^{(i)}$ for each rollout and computes the importance sampling ratio $r_{t}^{(i)}(\theta)$ between the updated policy and the behavior policy. 
We write $\ell_{t}^{(i)}(\theta) = \log r_{t}^{(i)}(\theta)$ as the log-importance ratio, which provides a local measure of policy deviation. 

We define the trust-region set as
\begin{equation}
\mathcal{I}
=
\big\{(i,t)\,:\,|\ell_{t}^{(i)}(\theta)| \le \delta \big\},
\label{eq:trust_region_set}
\end{equation}
which serves as a symmetric proxy for the unclipped region under PPO-style clipping.
Using this notation, EUR is defined as
\begin{equation}
\mathrm{EUR}
=
\frac{
\sum_{(i,t)\in\mathcal{I}} |\hat{A}_{t}^{(i)}|
}{
\sum_{i,t} |\hat{A}_{t}^{(i)}|
}.
\label{eq:eur_main}
\end{equation}

In Appendix~\ref{appendix:eur_proof}, we formally show that EUR provides a principled estimate of unclipped gradient contributions and serves as a proxy for controlling the upper bound of policy divergence. 
Consequently, a high EUR signifies stable and meaningful policy improvement, whereas a low EUR warns that the optimizer is effectively stalling due to suppressed gradients.

\textbf{Update Consistency (UC).}
\textbf{UC quantifies the variability of the unclipped updates, where larger values indicate greater inconsistency in their deviation magnitudes.}
Using the trust-region set $\mathcal{I}$ defined in Eq.~\eqref{eq:trust_region_set}, we compute the advantage-weighted mean log-ratio as
\[
\mu_\ell
=
\frac{
\sum_{(i,t)\in\mathcal{I}} |\hat{A}_{t}^{(i)}|\,\ell_{t}^{(i)}(\theta)
}{
\sum_{(i,t)\in\mathcal{I}} |\hat{A}_{t}^{(i)}|
},
\]
with this quantity in place, UC is defined as
\begin{equation}
\mathrm{UC}
=
\sqrt{
\frac{
\sum_{(i,t)\in\mathcal{I}}
|\hat{A}_{t}^{(i)}|\,\big(\ell_{t}^{(i)}(\theta) - \mu_\ell\big)^2
}{
\sum_{(i,t)\in\mathcal{I}} |\hat{A}_{t}^{(i)}|
}}.
\label{eq:uc_main}
\end{equation}

In Appendix~\ref{appendix:uc_proof}, we formally demonstrate that this metric is closely related to the variance of the local KL divergence.
Thus, UC provides a principled indicator of update stability within the trust region, allowing us to distinguish coherent policy improvements from noisy, destabilizing steps.

\textbf{Affinity.}
\textbf{Affinity quantifies the joint quality of policy optimization by synthesizing the volume of effective updates with the stability of their deviation magnitudes.}
Effective training requires balancing the quantity of unclipped updates against the variance of their divergence, as neither EUR nor UC is sufficient in isolation.
Using the trust-region threshold $\delta$ from Eq.~\eqref{eq:trust_region_set} and setting a temperature parameter $\tau = \delta/2$, we define
\begin{equation}
\textit{Affinity}
=
\mathrm{EUR}\cdot \exp\!\Big(-\frac{\mathrm{UC}}{\tau}\Big).
\label{eq:affinity_main}
\end{equation}
This formulation modulates the update volume EUR with an exponential decay based on the consistency metric UC.
In Appendix~\ref{appendix:affinity_proof}, we provide further theoretical derivations for this composite design.
Consequently, \textit{Affinity} serves as a robust scalar indicator that yields a high score only when the optimization is both sufficiently active and stable, effectively filtering out updates that are either negligible in volume or excessively noisy.
\subsection{HINT: Helping Ineffective Rollouts Navigate Towards Effectiveness}

Incorporating off-policy data into on-policy RL requires maintaining high \textit{Affinity} to ensure that external guidance translates into stable and effective policy updates.
However, prior methods consistently suffer from low \textit{Affinity}, as their reliance on Answer-Hints creates severe distributional mismatches that destabilize the training process.

Formally, we distinguish between Answer-Hints, which expose complete or partial reference-solution content, and \textit{Meta-Hints}, which provide abstract strategic scaffolding without revealing the solution itself.
Drawing from cognitive psychology, research on feedback levels demonstrates that process-oriented guidance promotes deeper understanding and generalization, whereas task-level feedback often leads to superficial dependence~\citep{hattie2007power}.
Despite this theoretical consensus, prior exploration methods in RL predominantly rely on Answer-Hints, thereby failing to activate the intrinsic problem-solving capabilities of the model.
To improve \textit{Affinity}, we guide the model toward productive reasoning trajectories using Meta-Hints.

To explicitly leverage this improved \textit{Affinity} for stable optimization, simply augmenting the data is insufficient; we require an objective that dynamically adapts to the quality of each update.
To this end, we propose Affinity-Aware Policy Optimization (AAPO), which re-weights the objective using the group-level affinity score $\alpha_q$:
\begin{equation}
\mathcal{J}_{\text{AAPO}}(\theta) 
= 
\mathbb{E}_{q} \left[ 
    \text{sg}(\alpha_q)^\lambda \cdot \mathcal{J}_{\text{GRPO}}^{(q)}(\theta)
\right],
\label{eq:aapo_loss}
\end{equation}
where $\mathcal{J}_{\text{GRPO}}^{(q)}(\theta)$ is the standard objective term defined in Eq.~\eqref{grpo}.
For each prompt group $q$, we compute group-level metrics $\mathrm{EUR}_q$ and $\mathrm{UC}_q$ over the corresponding trajectories and define
\[
\alpha_q = \mathrm{EUR}_q \cdot \exp\!\left(-\frac{\mathrm{UC}_q}{\tau}\right),
\]
which is the group-level form of Eq.~\eqref{eq:affinity_main}.
Crucially, we apply the stop-gradient operator $\text{sg}(\cdot)$ to ensure that $\alpha_q$ functions strictly as a scalar coefficient, which prevents the model from maximizing the objective by freezing parameters to artificially boost stability.
The hyperparameter $\lambda \ge 1$ acts as a sensitivity coefficient, which suppresses the gradient contribution from unstable updates characterized by low affinity scores while preserving high-quality learning signals.

Formally, as illustrated in Figure~\ref{fig:framework}, the HINT framework operates as an adaptive two-stage process that dynamically selects the optimization objective based on rollout outcomes.
In the first stage, for a problem $q$, the model samples a set of trajectories $\{o_{1}, \ldots, o_{G}\}$ which are evaluated to obtain rewards $\{r_{1}, \ldots, r_{G}\}$.
If these rewards are non-sparse (i.e., at least one is correct), the data is treated as on-policy, and we update the model using the standard GRPO objective.
Conversely, if the initial rewards are sparse, we activate the rescue stage by constructing a hint-augmented query $q_{h}$ with a Meta-Hint $h$ to resample a new set of trajectories $\{o^{h}_{1}, \ldots, o^{h}_{G}\}$.
To mitigate the potential instability introduced by this external guidance, we optimize these regenerated trajectories using the $\mathcal{J}_{\text{AAPO}}$ objective defined in Eq.~\eqref{eq:aapo_loss}.

Crucially, while $q_{h}$ guides the rollout, the gradient is computed against the original query $q$, ensuring the model learns to solve the task independently without relying on hints as input features.

\section{Experiments}
\subsection{Setup}
\label{sec:setup}
\textbf{Experimental Setup.}
Our experiments are conducted using Qwen2.5-7B, Qwen2.5-3B~\citep{team2024qwen2} and LLaMa3.1-8B~\citep{dubey2024llama} as backbone models. 
To ensure a fair and controlled comparison, we constructed a high-quality training set derived from the DAPO-Math-17K dataset~\citep{yu2025dapo}. 
This process involved using Qwen2.5-72B-Instruct~\citep{team2024qwen2} to generate four distinct reasoning trajectories for each problem. 
These outputs were then validated for correctness with Math Verify\footnote{https://github.com/huggingface/Math-Verify}, from which we retained 10k fully correct samples to form our final training data. 
For baseline methods that require a ground-truth reference solution, we designated the shortest of the four correct trajectories for each problem.

\textbf{Benchmarks.} 
We evaluate the generalization ability of HINT on seven datasets, covering both in-distribution and out-of-distribution scenarios, without using any hint during evaluation.
For mathematical reasoning, we adopt AIME24\footnote{https://huggingface.co/datasets/math-ai/aime24}, MATH-500~\citep{hendrycks2021measuring}, OlympiadBench~\citep{he2024olympiadbench}, and Minerva~\citep{lewkowycz2022solving}, which are widely used benchmarks. Since the test sets of AIME24 are relatively small, we report avg@32, while for the other datasets we use pass@1.
To assess complex reasoning and out-of-distribution generalization, we further evaluate on ARC-Challenge~\citep{clark2018think}, GPQA-Diamond~\citep{rein2024gpqa}, and MMLU-Pro~\citep{wang2024mmlu}. 
These benchmarks allow us to assess both in-distribution performance and out-of-distribution generalization. 

\begin{table*}[ht]
\caption{Main Performance Comparison of HINT against Baselines. HINT demonstrates significant performance gains on in-distribution datasets, improving the Qwen2.5-7B, Qwen2.5-3B, and LLaMa3.1-8B models by \textbf{14.5\%}, \textbf{17.7\%}, and \textbf{9.1\%} in average accuracy, respectively. Furthermore, \textbf{the method consistently outperforms baselines on out-of-distribution data, highlighting its strong generalization capabilities.}}
\centering
\small
\setlength{\tabcolsep}{9pt}
\resizebox{\linewidth}{!}{
\begin{tabular}{l BBBB R ddd R}
\toprule
\multirow{2}{*}{\cellcolor{white}\textbf{Methods}} &
\multicolumn{4}{c}{\cellcolor{white}\textbf{In-Distribution}} &
\multirow{2}{*}{\cellcolor{white}\textbf{Avg}} &
\multicolumn{3}{c}{\cellcolor{white}\textbf{Out-of-Distribution}} &
\multirow{2}{*}{\cellcolor{white}\textbf{Avg}} \\
\cmidrule(lr){2-5} \cmidrule(lr){7-9}
& \cellcolor{white}{AIME24} & \cellcolor{white}{Math} & \cellcolor{white}{Olympiad} & \cellcolor{white}{Minerva} & \cellcolor{white}{} & \cellcolor{white}{ARC} & \cellcolor{white}{GPQA} & \cellcolor{white}{MMLU}\\
\midrule
\rowcolor{bgcolor} \multicolumn{10}{c}{\textbf{Qwen2.5-7B}} \\
Vanilla & 9.8 & 50.2 & 34.0 & 19.5 & 28.4 & 85.3 & 25.6 & 46.0 & 52.3 \\
SFT & 11.2 & 72.8 & 36.2 & 28.8 & 37.3 & 85.1 & 25.6 & 46.2 & 52.3 \\
\hdashline
LUFFY (NIPS'25) & 13.4 & 77.0 & 38.6 & \underline{34.2} & 40.8 & 86.0 & 26.8 & 48.8 & 53.9 \\
BREAD (ICML'25) & 14.0 & 77.4 & 38.0 & 31.0 & 39.9 & 88.2 & 30.2 & \underline{49.3} & 55.9 \\
\hdashline
GRPO & 13.9 & 76.8 & 38.0 & 31.0 & 39.9 & 88.0 & 29.4 & 48.0 & 55.1 \\
GRPO + Meta-Hints & \underline{14.4} & \underline{79.6} & \underline{40.2} & 34.0 & \underline{42.1} & \underline{88.8} & \underline{30.4} & \textbf{50.2} & \underline{56.5} \\
\textbf{HINT (Ours)} & \textbf{14.6} & \textbf{80.4} & \textbf{42.2} & \textbf{34.4} & \textbf{42.9} & \textbf{89.0} & \underline{32.8} & \textbf{50.2} & \textbf{57.3} \\
\midrule
\rowcolor{bgcolor} \multicolumn{10}{c}{\textbf{Qwen2.5-3B}} \\
Vanilla & 2.9 & 39.8 & 12.0 & 9.8 & 16.1 & 44.8 & 11.4 & 28.8 & 28.3 \\
SFT & 5.3 & 54.8 & 20.6 & 19.6 & 25.1 & 46.4 & 11.0 & 32.0 & 29.8 \\
\hdashline
LUFFY (NIPS'25) & 5.8 & 62.2 & 29.6 & 22.2 & 30.0 & 70.2 & 15.2 & 34.2 & 39.9 \\
BREAD (ICML'25) & 6.3 & 62.0 & 29.0 & 24.4 & 30.4 & 72.0 & \underline{18.2} & \textbf{36.3} & 42.2 \\
\hdashline
GRPO & 6.0 & 60.4 & 26.0 & 23.6 & 29.0 & 74.4 & 16.0 & \underline{36.2} & 42.2 \\
GRPO + Meta-Hints & \underline{6.8} & \underline{66.4} & \underline{30.4} & \underline{25.2} & \underline{32.2} & \underline{77.6} & 18.0 & 35.0 & \underline{43.5} \\
\textbf{HINT (Ours)} & \textbf{7.4} & \textbf{68.8} & \textbf{32.8} & \textbf{26.0} & \textbf{33.8} & \textbf{78.8} & \textbf{20.4} & 35.5 & \textbf{44.9} \\
\midrule
\rowcolor{bgcolor} \multicolumn{10}{c}{\textbf{LLaMa3.1-8B}} \\
Vanilla & 0.0 & 9.4 & 2.1 & 3.2 & 3.7 & 0.0 & 0.0 & 0.0 & 0.0 \\
SFT & 0.2 & 14.4 & 4.4 & 8.4 & 6.9 & 52.4 & 18.3 & 26.5 & 32.4 \\
\hdashline
LUFFY (NIPS'25) & 0.5 & 25.2 & \textbf{7.4} & 14.4 & 11.9 & 66.8 & 25.5 & 33.3 & 41.9 \\
BREAD (ICML'25) & \underline{0.7} & 23.0 & \underline{7.0} & \textbf{16.6} & 11.8 & 70.4 & 27.2 & 33.9 & 43.8 \\
\hdashline
GRPO & 0.5 & 23.2 & 6.3 & 12.2 & 10.6 & 70.0 & 26.4 & 33.0 & 43.1 \\
GRPO + Meta-Hints & 0.5 & \underline{26.8} & 6.6 & 14.4 & \underline{12.1} & \underline{74.8} & \underline{28.0} & \underline{36.4} & \underline{46.4} \\
\textbf{HINT (Ours)} & \textbf{1.0} & \textbf{28.0} & \underline{7.0} & \underline{15.2} & \textbf{12.8} & \textbf{75.3} & \textbf{30.4} & \textbf{39.0} & \textbf{48.2} \\
\bottomrule
\end{tabular}
}
\label{tab:main_results}
\end{table*}

\textbf{Baselines.} 
We compare HINT against several existing methods, including: 
(1)\textbf{GRPO}~\citep{guo2025deepseek}: The vanilla Group Relative Policy Optimization algorithm.
(2)\textbf{SFT}: Standard Supervised Fine-Tuning.
(3)\textbf{LUFFY}~\citep{yan2025learning}: A hybrid approach that combines on-policy and off-policy training, ensuring that each sampled batch contains at least one correct trajectory.
(4)\textbf{BREAD}~\citep{zhang2025bread}: A binary search–based method that identifies a hint length such that the model’s rollouts are neither all correct nor all incorrect, and uses this balanced point as the hint for training. Further experimental details can be found in Appendix ~\ref{sec:experimental details} for full reproducibility.

\subsection{Main results}

Table~\ref{tab:main_results} presents a comprehensive comparison of HINT against several mainstream baselines, encompassing two Answer-Hints methods.
Overall, HINT demonstrates remarkable effectiveness across all model scales, improving the average in-distribution accuracy of the corresponding vanilla backbones by \textbf{14.5}, \textbf{17.7}, and \textbf{9.1} absolute points for Qwen2.5-7B, Qwen2.5-3B, and LLaMa3.1-8B, respectively.
Our detailed analysis reveals three key findings regarding the efficacy of our data strategy, the necessity of our optimization objective, and the generalization capabilities of our method.

\textbf{Meta-Hints foster genuine reasoning over answer memorization.}
First, our results demonstrate that process-oriented guidance is more effective than answer-centric supervision.
Across almost all benchmarks, the \textit{GRPO + Meta-Hints} variant consistently outperforms strong baselines like LUFFY and BREAD, which rely on Answer-Hints.
This performance gap suggests that Meta-Hints provide more useful guidance than Answer-Hints, which can encourage the model to follow rigid solution paths.
By constraining the reasoning space rather than dictating the exact solution, Meta-Hints make sparse-reward failures more likely to become informative learning opportunities.
  

\textbf{AAPO is essential for fully exploiting off-policy guidance.}
Second, the consistent gains from \textit{GRPO + Meta-Hints} to the full HINT framework show that better guidance alone is not enough. 
Across all three backbones, HINT improves over \textit{GRPO + Meta-Hints} on both in-distribution and out-of-distribution averages; for Qwen2.5-7B, for example, the scores rise from 42.1\% to 42.9\% and from 56.5\% to 57.3\%, respectively. 
This pattern indicates that hint-augmented rollouts still introduce off-policy noise that standard GRPO cannot fully absorb. 
AAPO addresses this issue by gating gradients with the affinity score $\alpha_q$, allowing the model to retain beneficial hint-induced updates while suppressing unstable ones.

\textbf{HINT activates generalized reasoning capabilities beyond mathematical memorization.} 
Finally, HINT exhibits robust out-of-distribution (OOD) generalization, suggesting that its benefits extend beyond the mathematical domain used for training. 
On ARC, GPQA, and MMLU, HINT achieves the best average OOD performance across all three backbones, with the largest gain appearing on LLaMa3.1-8B, whose average rises from 43.1\% under GRPO to 48.2\% under HINT.
These results suggest that HINT improves reasoning strategies that transfer to unseen domains, rather than only strengthening task-specific memorization.
\subsection{Training Dynamics}
\label{sec:affinity_analysis}
\begin{figure}[h]
  \centering
  \begin{subfigure}[b]{0.49\linewidth}
    \centering
    \includegraphics[width=\linewidth]{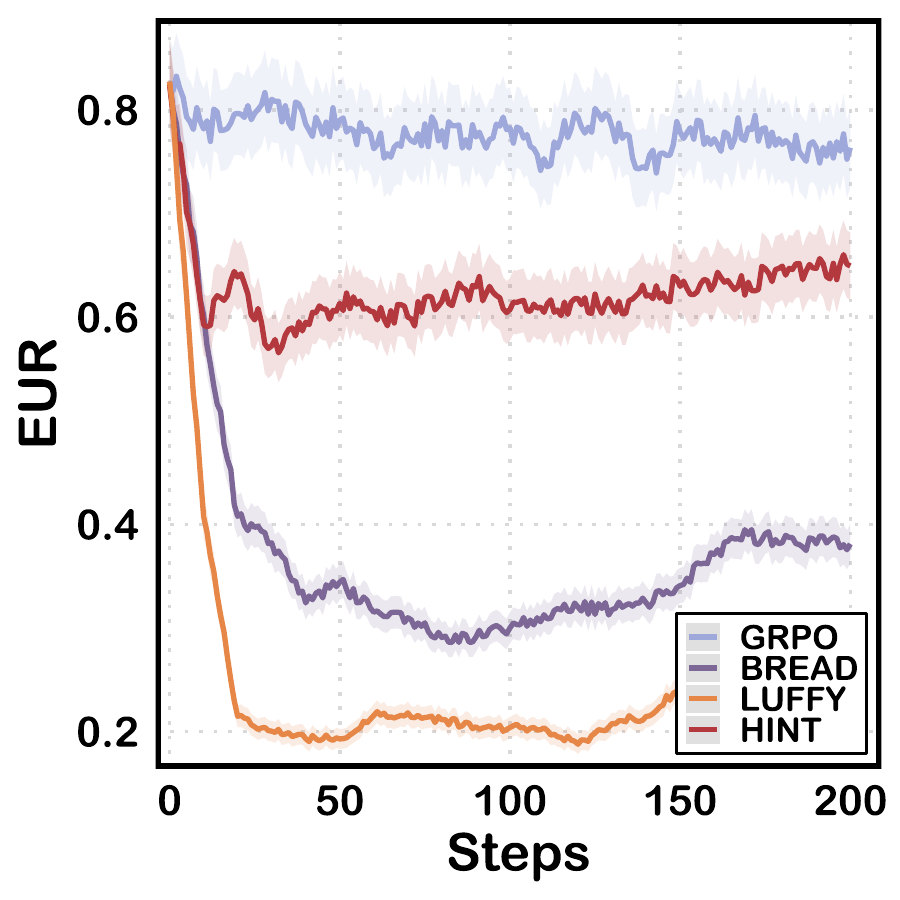} 
    \caption{EUR Metric}
    \label{fig:eur}
  \end{subfigure}
  \hfill
  \begin{subfigure}[b]{0.49\linewidth}
    \centering
    \includegraphics[width=\linewidth]{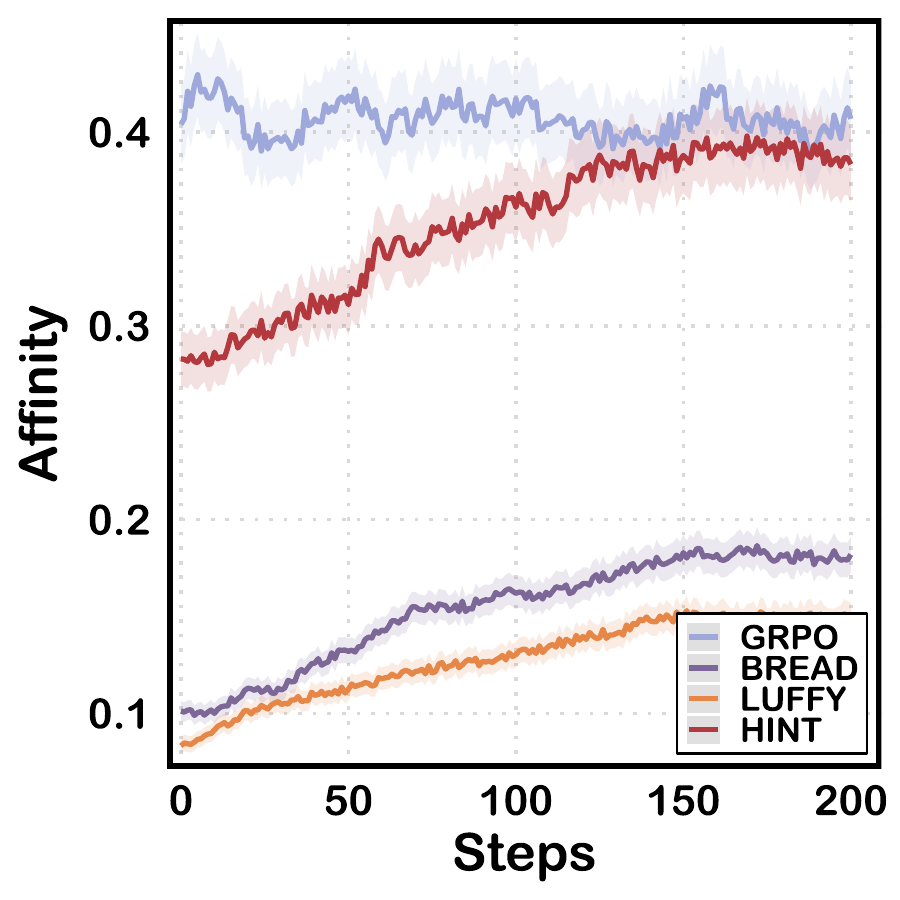}
    \caption{\textit{Affinity} Metric}
    \label{fig:affinity}
  \end{subfigure}
  \caption{Comparative analysis of training dynamics. 
  \textbf{(a)} HINT maintains a consistently high EUR, preventing the collapse seen in baselines. 
  \textbf{(b)} Consequently, HINT achieves significantly higher \textit{Affinity}.}
  \label{fig:eur_affinity}
\end{figure}
To investigate the impact of various strategies on training stability, we tracked the EUR and \textit{Affinity} metrics throughout the training process. The comparative results are plotted in Figure~\ref{fig:eur_affinity}, while the analysis of UC is detailed in Appendix~\ref{app:uc_analysis}.

\textbf{HINT prevents the ``EUR Collapse'' typical of off-policy learning.}
As illustrated in Figure~\ref{fig:eur}, traditional off-policy methods suffer from a severe ``EUR Collapse'', where the EUR plummets to near 0.2, indicating excessive clipping and wasted samples.
In sharp contrast, HINT avoids this failure mode.
While there is a slight initial adjustment, HINT maintains a high steady-state EUR that is much closer to GRPO than to other off-policy methods using Answer-Hints.
This confirms that HINT successfully keeps the policy updates within the trust region, ensuring high sample efficiency.

\textbf{High \textit{Affinity} validates the effectiveness of Meta-Hints.}
As presented in Figure~\ref{fig:affinity}, HINT is the only off-policy method that achieves and sustains high \textit{Affinity} scores.
While other methods stagnate at low \textit{Affinity} levels due to severe distributional mismatches, the \textit{Affinity} of HINT steadily increases and tracks the GRPO baseline.
This pattern supports our core proposition that Meta-Hints are more compatible with the model's intrinsic distribution, allowing external guidance to be incorporated as supervision rather than treated as interference.

\subsection{Does hinting truly enhance sample efficiency?}
\begin{figure}[h]
    \centering
    \begin{subfigure}[b]{0.48\columnwidth}
        \centering
        \includegraphics[width=\linewidth]{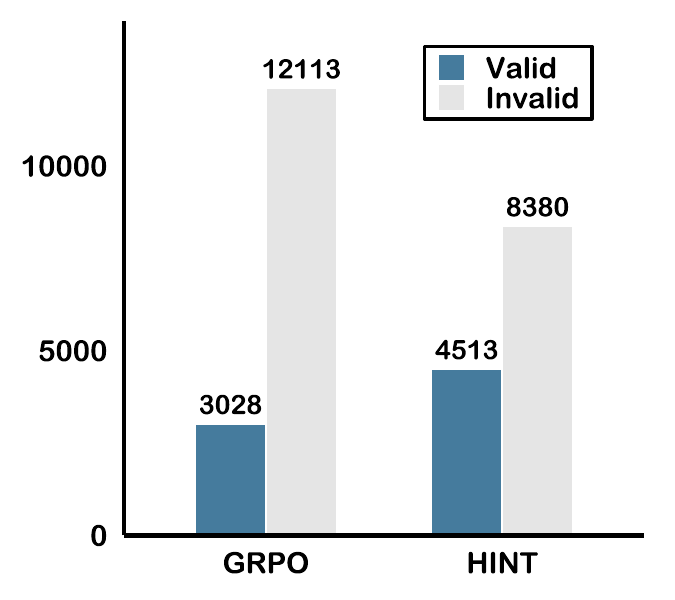}
        \caption{Throughput under fixed time budget}
        \label{fig:rollout_counts}
    \end{subfigure}
    \hfill
    \begin{subfigure}[b]{0.48\columnwidth}
        \centering
        \includegraphics[width=\linewidth]{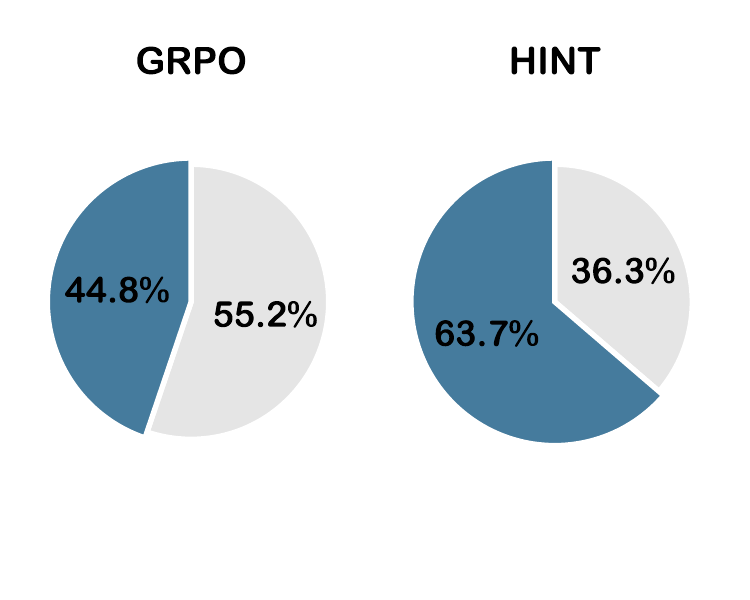}
        \caption{Validity rollout distribution over a full training epoch}
        \label{fig:validity_rates}
    \end{subfigure}
    \caption{Efficiency analysis. \textbf{(a)} HINT produces a higher net volume of valid trajectories despite lower total generation speed. \textbf{(b)} HINT significantly increases the proportion of effective training data.}
    \label{fig:valid_invalid_combined}
\end{figure}

\textbf{HINT significantly enhances both sampling efficiency and the density of effective supervision.}
To quantify this, we conducted a two-fold analysis evaluating generation speed under a fixed 8-hour budget and global data distribution over a full training epoch.
As illustrated in Figure~\ref{fig:rollout_counts}, although the inference overhead of Meta-Hints results in fewer total samples, HINT successfully yields a substantially higher volume of valid samples, defined as rollouts containing correct reasoning steps.
Specifically, it produces \textbf{1,485 more valid samples} than the standard GRPO baseline, indicating that the computational cost of generating hints is outweighed by the gain in exploration success.
Furthermore, Figure~\ref{fig:validity_rates} shows that, relative to the standard GRPO baseline, the validity rate increases from 44.8\% to 63.7\%, representing an absolute gain of \textbf{18.9\%}.
This shift indicates that HINT steers the model toward more productive regions of the solution space.
By reducing the prevalence of uninformative trajectories, HINT makes more of the collected data useful for optimization, thereby maximizing the utility of the available computational budget.

\subsection{Does external feedback affect generation diversity?}
\begin{table}[h]
  \centering
  \small 
  \setlength{\tabcolsep}{4pt} 
  \caption{Quantitative analysis of exploration diversity using average entropy. 
  \textbf{HINT promotes broader exploration compared to Answer-Hints.} 
  Here, ``w/ Off.'' denotes trajectories augmented with guidance, while ``w/o Off.'' refers to standard on-policy rollouts.}
  \begin{tabular}{lccc} 
  \toprule
    & w/ Off. & w/o Off. & All \\
  \midrule
  GRPO    &    --    & 0.143 & 0.143 \\
  LUFFY    &    --    & 0.174 & \underline{0.174} \\
  BREAD    &  \underline{0.128}  & \underline{0.183} & 0.162 \\
  HINT &  \textbf{0.188}  & \textbf{0.198} & \textbf{0.193} \\
  \bottomrule
  \end{tabular}
  \label{tab:entropy_comparison}
\end{table}

\textbf{HINT fosters broad and diverse exploration rather than converging to a narrow set of solutions.}
To quantify this, we analyzed the output distribution using average entropy as a metric for exploration breadth, with results detailed in Table~\ref{tab:entropy_comparison}.
Strategies relying on Answer-Hints, such as BREAD, exhibit the lowest entropy of 0.128 on the off-policy subset.
This indicates that providing explicit answers acts as a rigid constraint that narrows exploration toward a fixed path.
In contrast, HINT maintains a significantly higher entropy of 0.188 even under guidance, suggesting that abstract Meta-Hints guide the reasoning process without prescribing a single trajectory.
Crucially, this benefit extends to standard on-policy rollouts where HINT achieves the highest entropy of 0.198 among all methods.
Taken together, these results show that HINT preserves exploration diversity while still providing useful guidance.
\subsection{Does HINT scale with reasoning complexity?}
\begin{figure}[h]
  \centering
  \includegraphics[width=0.99\linewidth]{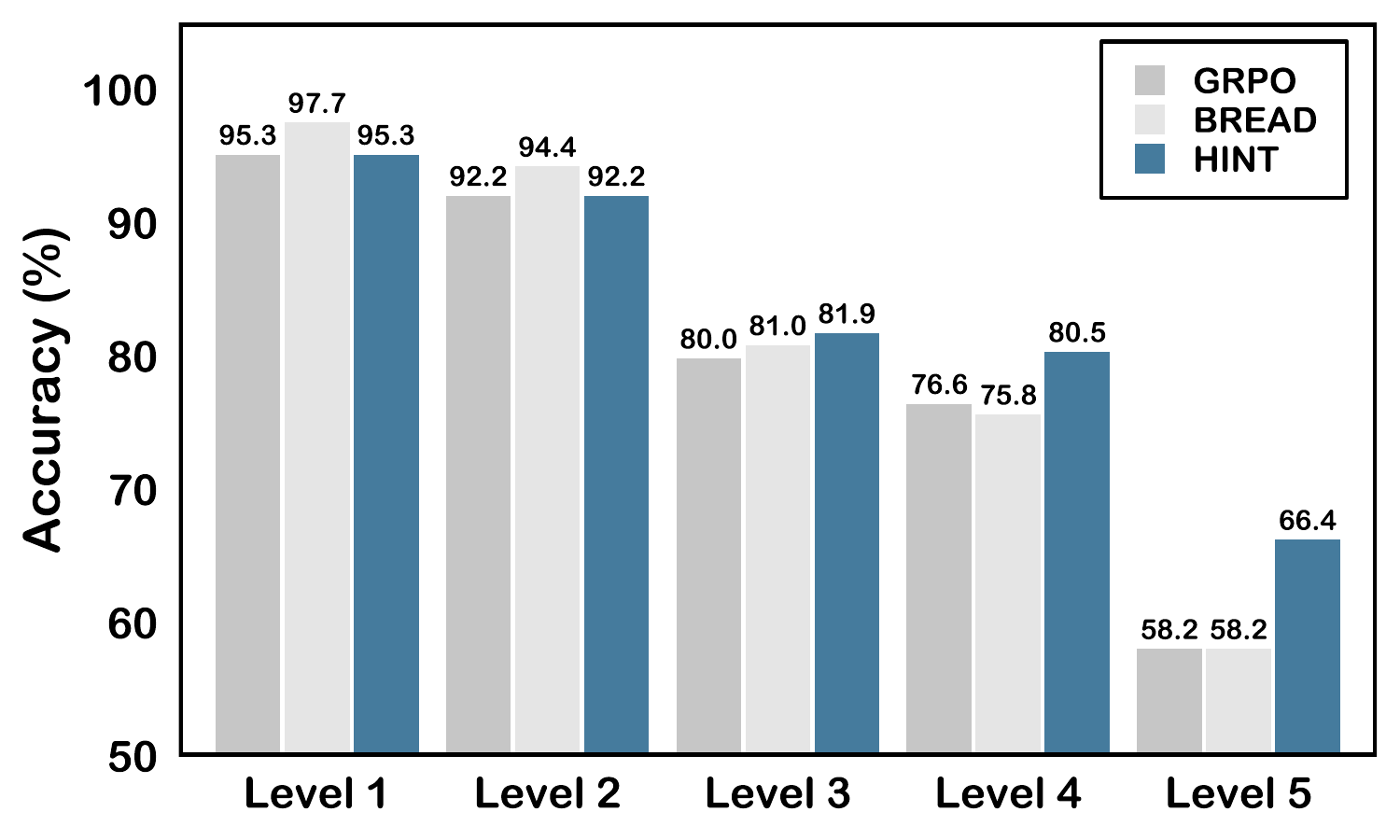}
  \caption{Performance comparison across different difficulty levels on the MATH-500 benchmark. While baseline methods plateau on hard tasks, HINT demonstrates \textbf{widening performance gaps as difficulty increases}, achieving an \textbf{8.2\% absolute gain} on Level 5 problems.}
  \label{fig:difficulty_breakdown}
\end{figure}

\textbf{HINT acts as a vital cognitive scaffold that specifically enhances performance on complex reasoning tasks.}
To verify this, we stratified the performance on the MATH-500 benchmark across five difficulty levels as illustrated in Figure~\ref{fig:difficulty_breakdown}.
On simpler tasks classified as Levels 1 and 2, all methods exhibit high competency with accuracy rates exceeding 92\%.
BREAD slightly outperforms the others in this regime.
This suggests that answer-level hints are already sufficient when the task is easy and the required reasoning pattern is simple.

However, a distinct performance divergence emerges as complexity increases.
On the most challenging Level 5 problems, both GRPO and BREAD stagnate at an identical accuracy of 58.2\%.
In contrast, HINT achieves a robust 66.4\% and marks a substantial absolute gain of 8.2\%.
This trend indicates that the benefits of Meta-Hints grow with task complexity.
While the intrinsic policy often suffices for simple scenarios, the strategic guidance of Meta-Hints becomes increasingly valuable as the search space expands.
This breakdown supports the claim that HINT is especially helpful for deep reasoning tasks.

\section{Conclusion}
We address the fundamental trade-off between exploration efficiency and update stability in RL for reasoning. 
We revealed that conventional Answer-Hints often induce low \textit{Affinity}, leading to unstable gradients despite high rewards. 
Our solution, HINT, resolves this conflict by combining Meta-Hints for high-level conceptual guidance with AAPO, a novel optimization objective that dynamically filters noise based on our proposed affinity metric. 
Empirical results confirm that HINT significantly outperforms strong baselines, particularly in scenarios requiring generalization beyond the training distribution. 
By providing a principled mechanism to leverage off-policy data without compromising stability, our work offers a robust foundation for training the next generation of reasoning models. 
Future directions include applying HINT to broader domains and exploring its synergy with iterative self-correction mechanisms.
\section{Limitations}
Despite the promising results, our work has several limitations that we plan to address in future research.
First, due to computational constraints, our experimental evaluation is primarily conducted on models with parameters ranging from 3B to 8B.
While HINT demonstrates consistent gains across these scales, its efficacy on significantly larger models (e.g., 70B or larger) remains to be empirically verified.
Second, the current implementation of HINT focuses exclusively on text-based reasoning tasks.
We have not yet explored its application to multimodal scenarios, such as visual mathematical problem solving, where integrating visual cues into the meta-hint generation process presents a unique challenge.
We leave the extension of our framework to larger-scale models and multimodal domains as directions for future work.

\section{Acknowledgments}
This work was supported by Ant Group.
\bibliography{ref}
\clearpage
\appendix
\section*{Appendix}
\section{Theoretical Foundations of EUR, UC, and Affinity}
\subsection{Proofs for EUR}
\label{appendix:eur_proof}

In this section, we provide the theoretical justification for the two main claims made in the main paper regarding the EUR:  
(I) EUR estimates the fraction of unclipped PPO gradient contributions \citep{schulman2017proximal};  
(II) EUR serves as a proxy for bounding policy divergence in the sense of TRPO’s monotonic improvement guarantee \citep{schulman2015trust}.

\subsubsection{Preliminaries}

For each token step \(i\), let  
\[
r_i = \frac{\pi_\theta(a_i \mid s_i)}{\pi_{\theta_{\mathrm{old}}}(a_i \mid s_i)}, 
\qquad  
\ell_i = \log r_i.
\]
PPO optimizes a clipped surrogate objective \citep{schulman2017proximal}, defined as
\begin{equation}
\small
L_{\mathrm{CLIP}}(\theta)
=
\hat{\mathbb{E}}_i\Big[
\min\big(r_i A_i,\;
\operatorname{clip}(r_i,1 \pm \varepsilon)\,A_i\big)
\Big],
\label{eq:ppo_clip_app}
\end{equation}
and then maximizes $L_{\mathrm{CLIP}}(\theta)$ with respect to $\theta$.

Let \(\mathcal{I}=\{i:|r_i-1|\le\varepsilon\}\) denote the set of unclipped updates and \(\mathcal{C}\) the clipped ones.  
The gradient of \eqref{eq:ppo_clip_app} decomposes as:
\begin{equation}
\label{eq:ppo_grad_decomp_app}
\nonumber
\begin{split}
\nabla_\theta L_{\mathrm{CLIP}} &= \mathbb{E}[\,\nabla_\theta(r_iA_i)\,\mathbf{1}(i\in\mathcal{I})\,] \\
&\quad + \mathbb{E}[\,\nabla_\theta(r^{\mathrm{clip}}_i A_i)\,\mathbf{1}(i\in\mathcal{C})\,].
\end{split}
\end{equation}

As noted in \citet{schulman2017proximal}, gradients from clipped terms either vanish or are directionally distorted, while terms in \(\mathcal{I}\) preserve the correct policy gradient direction.

The Effective Update Ratio is defined in the main paper as:
\begin{equation}
\nonumber
\mathrm{EUR}
=
\frac{\sum_i |A_i|\,\mathbf{1}(|\ell_i|\le\delta)}
{\sum_i |A_i|}.
\label{eq:eur_def_app}
\end{equation}

\subsubsection{Proof of Claim (i): EUR estimates the fraction of unclipped PPO gradient contributions}

We demonstrate that EUR provides a principled empirical estimate of the proportion of gradient contributions arising from unclipped PPO updates. 
Recall that, for token-level PPO, the unclipped surrogate gradient at position $i$, denoted as $g_i$, is given by:
\begin{equation}
\begin{aligned}
g_i &= \nabla_\theta(r_i A_i) \\
    &= A_i\, r_i\, \nabla_\theta \log \pi_\theta(a_i\mid s_i),
\end{aligned}
\label{eq:grad_deriv}
\end{equation}
where $r_i = \frac{\pi_\theta(a_i\mid s_i)}{\pi_{\theta_{\text{old}}}(a_i\mid s_i)}$. 
For updates within the trust region (i.e., $i\in\mathcal{I}$ with $|\ell_i|\le\delta$), we have $r_i = e^{\ell_i}\approx 1$ given that $\ell_i$ is small. 
Consequently, the gradient magnitude simplifies to:
\begin{equation}
\nonumber
\|g_i\| \approx |A_i| \,\|\nabla_\theta \log\pi_\theta(a_i\mid s_i)\|.
\end{equation}
Since $\|\nabla_\theta \log\pi_\theta(a_i\mid s_i)\|$ is locally bounded and relatively stable across nearby policy iterates, variations in $\|g_i\|$ are dominated by variations in $|A_i|$. Thus, the total contribution of unclipped updates to the gradient is proportional to:
\begin{equation}
\nonumber
\mathbb{E}\big[\,|A_i|\,\mathbf{1}(i\in\mathcal{I})\,\big].
\end{equation}
Similarly, the total gradient magnitude (including both clipped and unclipped updates) is proportional to $\mathbb{E}[\,|A_i|\,]$. 
Therefore, the fraction of gradient contributions originating from unclipped updates is:
\begin{equation}
\nonumber
\text{EUR} \approx \frac{
\mathbb{E}\big[\,|A_i|\,\mathbf{1}(i\in\mathcal{I})\,\big]
}{
\mathbb{E}\big[\,|A_i|\,\big]
}.
\end{equation}
By construction, this matches our definition of EUR, confirming it as an effective estimator for the fraction of gradient contributions unsuppressed by clipping.

\subsubsection{Proof of Claim (ii): EUR controls policy divergence in the TRPO sense}

TRPO~\citep{schulman2015trust} establishes a monotonic improvement lower bound dependent on the KL divergence:
\begin{equation}
\nonumber
\label{eq:trpo_bound_app}
\eta(\theta)
\;\ge\;
L_{\theta_{\mathrm{old}}}(\theta)
-
C \cdot D_{\mathrm{KL}}^{\max}(\pi_{\theta_{\mathrm{old}}},\pi_\theta),
\end{equation}
where \(C\) is a constant dependent on $\gamma$ and $\epsilon$.
The token-level empirical KL divergence can be approximated by the expectation of log-ratios:
\begin{equation}
\nonumber
D_{\mathrm{KL}}(\pi_{\theta_{\mathrm{old}}}\Vert\pi_\theta)
\approx
\mathbb{E}_{s,a\sim\pi_{\mathrm{old}}}\big[\,|\ell_i|\,\big].
\end{equation}

Recall that $\mathrm{EUR}$ is the advantage-weighted fraction of updates within the trust region ($|\ell_i| \le \delta$). Let $\mathcal{C} = \{i : |\ell_i| > \delta\}$ denote the set of clipped updates. The relationship between EUR and the probability mass of $\mathcal{C}$ depends on the distribution of advantages.

\vspace{5pt}
\noindent\textbf{Assumption 1.} \textit{The expected magnitude of advantages for clipped updates is lower bounded by a factor of the global expected magnitude, i.e., $\mathbb{E}[|A_i| \mid i \in \mathcal{C}] \ge \alpha \mathbb{E}[|A_i|]$ for some $\alpha > 0$.}
\vspace{5pt}

Under this mild assumption, we can relate EUR to the probability of clipping $P(\mathcal{C})$:
\begin{equation}
\begin{split}
\nonumber
1 - \mathrm{EUR} 
&= \frac{\sum_{i \in \mathcal{C}} |A_i|}{\sum_{\mathrm{all}} |A_i|} \\
&\approx \frac{P(\mathcal{C}) \cdot \mathbb{E}[|A_i| \mid \mathcal{C}]}{\mathbb{E}[|A_i|]} \\
&\ge \alpha P(\mathcal{C}).
\end{split}
\end{equation}
This implies $P(\mathcal{C}) \le \frac{1-\mathrm{EUR}}{\alpha}$. Conversely, the contribution to the KL divergence from clipped samples is lower bounded:
\begin{equation}
\nonumber
\begin{split}
D_{\mathrm{KL}} 
&\ge P(\mathcal{C}) \cdot \min_{i \in \mathcal{C}} |\ell_i| \\
&> P(\mathcal{C}) \cdot \delta.
\end{split}
\end{equation}
If EUR is low (close to 0), the advantage mass is concentrated in $\mathcal{C}$. Unless the advantages in $\mathcal{C}$ are negligibly small (which contradicts meaningful exploration), a low EUR implies a significant $P(\mathcal{C})$, forcing $D_{\mathrm{KL}}$ to exceed the trust region boundary $\delta$.
Therefore, maintaining a high EUR is a necessary proxy for constraining $D_{\mathrm{KL}}$ and preserving the validity of the TRPO bound.

\subsubsection{Summary}
Taken together, the results above show that EUR simultaneously quantifies the fraction of gradient mass preserved by the unclipped PPO surrogate and provides a practical handle on the policy divergence term appearing in TRPO’s monotonic improvement bound. 
Consequently, a high EUR indicates that most updates lie within a stable trust-region regime where policy gradients remain informative, whereas a low EUR reveals that clipped updates dominate the optimization process, leading to vanishing effective gradients and ineffective learning.

\subsection{Proofs for UC}
\label{appendix:uc_proof}

In this section, we provide the theoretical justification for the UC metric introduced in the main paper.  
We show that UC can be interpreted as 
(I) an advantage-weighted measure of variability in local log-importance ratios among unclipped updates, and 
(II) a proxy for the variance of the local KL divergence, which is closely tied to the stability of policy updates.

\subsubsection{Preliminaries}

Recall that for each token step \(i\), we define
\[
r_i
=
\frac{\pi_\theta(a_i \mid s_i)}{\pi_{\theta_{\mathrm{old}}}(a_i \mid s_i)},
\qquad
\ell_i = \log r_i,
\]
and the trust-region condition \(|\ell_i|\le\delta\) identifies the set of unclipped updates:
\[
\mathcal{I} = \{\, i : |\ell_i|\le\delta \,\}.
\]
The token-level advantages are denoted by \(A_i\), and we use the absolute values \(|A_i|\) as
importance weights on the contribution of each token.

Within the set \(\mathcal{I}\), we define the advantage-weighted mean log-ratio:
\begin{equation}
\nonumber
\mu_\ell 
=
\frac{\sum_{i\in\mathcal{I}} |A_i| \,\ell_i}{\sum_{i\in\mathcal{I}} |A_i|},
\label{eq:uc_mean_logratio_app}
\end{equation}
and the UC is given by the advantage-weighted standard deviation:
\begin{equation}
\mathrm{UC}
=
\sqrt{
\frac{
\sum_{i\in\mathcal{I}} |A_i|\,(\ell_i - \mu_\ell)^2
}{
\sum_{i\in\mathcal{I}} |A_i|
}
}.
\label{eq:uc_def_app}
\end{equation}

\subsubsection{UC as a measure of variability among effective updates}

As shown in ~\eqref{eq:uc_def_app}, UC is precisely the standard deviation of the log-importance ratios $\ell_i$ over the set of effective updates $\mathcal{I}$. 
A small UC indicates that the $\ell_i$ values within $\mathcal{I}$ are tightly concentrated around their weighted mean $\mu_\ell$, implying that the magnitudes of the effective updates are consistent and that the resulting policy changes are approximately uniform across token positions. 
In contrast, a large UC reflects substantial variability among the $\ell_i$ values: some effective updates correspond to very small log-ratios (i.e., conservative steps), while others lie close to the trust-region boundary (i.e., aggressive steps). 
Such heterogeneity results in uneven and potentially unstable policy updates.

Formally, define the normalized weights
\[
\tilde{w}_i = \frac{|A_i|}{\sum_{j\in\mathcal{I}} |A_j|}, \quad i\in\mathcal{I}.
\]
Then \eqref{eq:uc_def_app} can be rewritten as
\[
\mathrm{UC}^2 
= 
\sum_{i\in\mathcal{I}} \tilde{w}_i (\ell_i - \mu_\ell)^2,
\]
which is the weighted variance of \(\ell_i\) under the empirical distribution induced by the
advantages \(|A_i|\).  
Thus UC quantifies how “spread out” the log-ratios are among those updates that are not clipped.

\subsubsection{Relation between UC and gradient variance}

We now connect UC to the variance of the policy gradient updates. Consider the gradient contribution magnitude for a single token $i$ within the trust region ($i \in \mathcal{I}$), defined as $X_i = A_i r_i \approx A_i(1+\ell_i)$.
The stability of training depends on the variance of this update scale. Assuming that the advantage $A_i$ and the log-ratio $\ell_i$ are uncorrelated within the local trust region, we evaluate $\mathrm{Var}(X_i)$ using the standard variance decomposition approximation:
\begin{equation}
\nonumber
\begin{split}
\mathrm{Var}(g_i) &\propto \mathrm{Var}\big(A_i(1+\ell_i)\big) \\
&\approx \mathrm{Var}(A_i) + \mathrm{Var}(A_i \ell_i).
\end{split}
\end{equation}
The first term, $\mathrm{Var}(A_i)$, represents the inherent variance of the reward structure (baseline variance), which is irreducible by policy constraints. The second term captures the variance introduced by the policy shift. Applying the product variance decomposition to $A_i \ell_i$:
\begin{equation}
\label{eq:uc_variance_decomp}
\begin{split}
\mathrm{Var}(A_i \ell_i) \approx \;& \mathbb{E}[A_i^2] \mathrm{Var}(\ell_i) \\
&+ \mathbb{E}[\ell_i]^2 \mathrm{Var}(A_i).
\end{split}
\end{equation}
Inside the trust region, $\ell_i$ is centered near 0, making the term $\mathbb{E}[\ell_i]^2$ negligible. Thus, the dominant component of the induced variance simplifies to:
\begin{equation}
\nonumber
\mathrm{Var}_{\mathrm{induced}} \approx \mathbb{E}[A_i^2] \cdot \mathrm{Var}(\ell_i).
\end{equation}
Recall that $\mathrm{UC}^2$ is defined as the advantage-weighted variance of $\ell_i$. Although strictly distinct from the unweighted $\mathrm{Var}(\ell_i)$, they are empirically aligned.
As shown in \eqref{eq:uc_variance_decomp}, UC acts as a multiplicative gain on the gradient variance. A high UC amplifies the gradient noise proportional to the squared advantages $\mathbb{E}[A_i^2]$, thereby destabilizing the update direction.
Consequently, minimizing UC is theoretically justified to dampen the variance of policy updates specifically arising from diverse importance ratios.

\subsubsection{Relation between UC and local KL variability}

We next relate UC to the variability in local KL divergence. The per-state KL divergence between the old and new policy can be expressed as:
\begin{equation}
\nonumber
\begin{split}
D_{\mathrm{KL}}\big(\pi_{\theta_{\mathrm{old}}}(\cdot\mid s) \;&\Vert\; \pi_{\theta}(\cdot\mid s)\big) \\
&= \mathbb{E}_{a\sim\pi_{\theta_{\mathrm{old}}}(\cdot\mid s)}\big[\log r(a,s)\big].
\end{split}
\end{equation}
At the token level, the empirical KL is estimated by averaging \(\ell_i\) over samples from \(\pi_{\theta_{\mathrm{old}}}\). Thus, the variability of \(\ell_i\) within \(\mathcal{I}\) directly reflects how much the local per-state KL fluctuates around its mean.

Since the monotonic improvement bound of TRPO~\citep{schulman2015trust} relies on controlling the KL divergence, large fluctuations in \(\ell_i\) (i.e., a high UC) suggest that certain states experience near-boundary policy shifts, even if the average KL remains small. This phenomenon effectively weakens the trust-region assumption and may induce oscillatory learning dynamics. By contrast, a low UC ensures that per-token KL changes are not only small on average but also uniformly bounded, leading to more reliable surrogate optimization.

\subsubsection{Summary}

In summary, UC captures the internal stability of policy updates within the trust region by measuring the advantage-weighted variance of log-importance ratios among unclipped samples. A low UC implies that effective updates move the policy in a coherent and conservative manner, whereas a high UC reveals that updates, though nominally ``valid,'' are heterogeneous and prone to inducing instability. Together with EUR, UC provides a complementary view of both the quantity and the quality of effective policy updates during training.

\subsection{Theoretical Discussion of Affinity}
\label{appendix:affinity_proof}

In this section, we provide the theoretical motivation for combining EUR and UC into the unified \textit{Affinity} metric introduced in the main paper. We demonstrate that \textit{Affinity} captures the joint requirements for effective and stable policy updates in PPO-style RL and relate its formulation to the principles underlying trust-region optimization.

\subsubsection{Preliminaries}

We briefly recall the definitions of EUR and UC. Let \(\ell_i = \log\frac{\pi_\theta(a_i\mid s_i)}{\pi_{\theta_\mathrm{old}}(a_i\mid s_i)}\) denote the log-importance ratio at token step \(i\), and let \(\mathcal{I} = \{ i: |\ell_i|\le \delta \}\) be the set of unclipped updates. EUR measures the fraction of effective updates:
\begin{equation}
\nonumber
\mathrm{EUR} = \frac{\sum_{i}|A_i|\mathbf{1}(i\in\mathcal{I})}{\sum_{i}|A_i|}.
\end{equation}
UC quantifies the internal variability of those updates. Defined formally:
\begin{align}
\nonumber
\mathrm{UC} &= \sqrt{\frac{\sum_{i\in\mathcal{I}} |A_i|(\ell_i-\mu_\ell)^2}{\sum_{i\in\mathcal{I}} |A_i|}}, \\
\nonumber
\mu_\ell &= \frac{\sum_{i\in\mathcal{I}} |A_i|\ell_i}{\sum_{i\in\mathcal{I}} |A_i|}.
\end{align}

\subsubsection{Rationale for combining EUR and UC}

As shown in Appendix~\ref{appendix:eur_proof}, EUR provides a principled empirical estimate of the proportion of gradient mass preserved by the unclipped PPO surrogate. Hence, a high EUR indicates that most updates meaningfully contribute to the policy gradient. However, EUR alone cannot ensure stability: if the log-ratios within \(\mathcal{I}\) vary widely (high UC), many of those ``effective'' updates may be close to the trust-region boundary, potentially inducing oscillatory policy shifts.

Appendix~\ref{appendix:uc_proof} further demonstrates that UC approximates the variance of token-level policy divergence, characterizing the consistency of unclipped gradients. Yet, UC alone is insufficient: a perfectly consistent set of updates (low UC) yields little value if EUR is small, as most gradients would be clipped, resulting in negligible policy movement.

Therefore, a high-quality update requires satisfying both conditions simultaneously: a sufficiently large proportion of effective updates (high EUR) and low variability among them (low UC).

\subsubsection{Affinity as a joint stability-efficiency indicator}

To encode this joint requirement into a single scalar, we define the \textit{Affinity} metric:
\begin{equation}
\nonumber
\label{eq:affinity_appendix}
\mathrm{Affinity} = \mathrm{EUR}\cdot \exp\!\left(-\frac{\mathrm{UC}}{\tau}\right), \quad \tau = \frac{\delta}{2}.
\end{equation}
This multiplicative formulation is motivated by two key factors:

\vspace{5pt}
\noindent\textbf{Logical conjunction.}
The product structure ensures that a failure in either condition (low EUR or high UC) produces a proportionally low \textit{Affinity}. This captures the fact that effective PPO-style updates necessitate the simultaneous satisfaction of both conditions.

\vspace{5pt}
\noindent\textbf{Exponential penalty on inconsistency.}
Since UC measures the weighted variance in log-ratios, the term \(\exp(-\mathrm{UC}/\tau)\) acts analogously to an inverse smoothness regularizer, sharply penalizing updates near the trust-region boundary. The temperature term \(\tau=\delta/2\) scales the penalty, ensuring it becomes substantial when UC approaches the limit of the trust region.

\subsubsection{Relationship to trust-region optimization}

Trust-region methods (including TRPO) rely on bounding the KL divergence to guarantee monotonic policy improvement. While EUR controls the fraction of updates satisfying the trust-region condition (reflecting the mean KL contribution), UC characterizes the variability of the local KL divergence within that region. Consequently, \textit{Affinity} integrates both aspects: high \textit{Affinity} indicates that the empirical KL is not only small (ensured by high EUR) but also stable across updates (ensured by low UC), aligning with the conditions under which trust-region guarantees are most effective.

\subsubsection{Summary}

\textit{Affinity} synthesizes two complementary perspectives on PPO update quality: \textbf{(I) the proportion of effective updates (EUR)}, and \textbf{(II) the consistency of those updates (UC)}. The multiplicative formulation in \eqref{eq:affinity_appendix} captures the synergy required for reliable policy improvement, providing a practical scalar diagnostic for monitoring exploration efficiency and training stability.

\clearpage

\section{Experimental Details}
\label{sec:experimental details}

\subsection{Detailed Setup}
\label{sec:detailed setup}
\textbf{Platform.} All of our
experiments are conducted on workstations equipped with 8 NVIDIA A100 PCIe GPUs with
80GB memory.

\textbf{Training Data.}
The training was performed using a carefully selected subset of the DAPO-Math-17K dataset~\citep{yu2025dapo}. 
As the original dataset lacks ground-truth solutions, we curated our own by first using Qwen2.5-72B-Instruct to generate four reasoning trajectories for each problem. 
After validating the final answers with \textit{Math-verify}, we compiled a high-quality training set of 10k problems for which all four generated trajectories were correct. 
For baselines requiring a ground truth, the most token-efficient of these four correct trajectories was designated as the ground truth.
For our methods, we pre-generated the required heuristic hints for the entire 10k-sample training set using Qwen2.5-72B-Instruct.
The prompts used in the above process will be detailed in Section~\ref{sec:prompt list}.

\textbf{Important Parameters of HINT. } 
HINT is implemented based on the open-source Rl framework lsrl\footnote{https://github.com/lsdefine/lsrl}. 
The RL algorithm employs the GRPO advantage estimator with no KL penalty (kl\_coef is set to 0.0). 
The clipping parameter $\epsilon$ is set to {0.2}. 
For each group, 8 answers are generated, and the training batch size is set to 2. 
Distributed training utilizes the DeepSpeed library with the \textit{AdamW} optimizer and a learning rate of 1e-6. 
The \textit{train batch size} is set to 8, \textit{gen batch size} is set to 32, \textit{accum steps} is set to 64, \textit{gen update steps} is set to 128, \textit{temperature} is set to 0.9, \textit{max response} is set to 8192. 
Mixed-precision training with BF16 is enabled. 
Memory optimization employs ZeRO Stage 2, with optimizer state offloading to CPU. 

\textbf{Important Parameters of Other Baselines.}
For baselines with publicly available code repositories, we utilized their official implementations and the parameters specified in their respective publications. 
For methods without public code, such as BREAD\citep{zhang2025bread} and QuestA\citep{li2025questa}, we reproduced their results using the lsrl framework, strictly adhering to the experimental parameters detailed in their papers.

\textbf{Reward Setup.}  
For our experiments, we employ a sparse, binary reward function. The reward is determined exclusively by the correctness of the final answer in a model's generated trajectory. We use the \textit{Math-Verify} tool for automatic verification, assigning a reward of \textbf{+1} for a correct final answer and \textbf{0} for an incorrect one.

\subsection{Prompt List}
\label{sec:prompt list}
\begin{tcolorbox}
[title=Prompt Template for GRPO, colback=white, colframe=black!75!white, breakable]
\linespread{1.3}\selectfont
\textbf{System:} You are a helpful AI assistant. A conversation takes place between the User and the Assistant. The User asks a question, and the Assistant solves it. Please help me solve this question. Wrap only the final answer in \textit{\textbackslash \textbackslash boxed\{\}}. \vspace{1em}

\textbf{Question:} [Question] \vspace{1em}

\textbf{User:}
\end{tcolorbox}

\vspace{1em}  

\begin{tcolorbox}
[title=Prompt Template for HINT, colback=white, colframe=black!75!white, breakable]
\linespread{1.3}\selectfont
\textbf{System:} You are a helpful AI assistant. A conversation takes place between the User and the Assistant. The User asks a question, and the Assistant solves it. Please help me solve this question. Wrap only the final answer in \textit{\textbackslash \textbackslash boxed\{\}}. \vspace{1em}

\textbf{\textcolor{red}{Hint:}} Here are some key information provided to assist you in solving the problem: [Hint] \vspace{1em}

\textbf{Question:} [Question] \vspace{1em}

\textbf{User:}
\end{tcolorbox}

\vspace{1em} 

\begin{tcolorbox}
[title=Prompt Template for Generating hints, colback=white, colframe=black!75!white, breakable]
\linespread{1.3}\selectfont
\textbf{System:} 

* Role and Goal

You are a top-tier problem-solving expert and a master educator. 
Your goal is not to solve the problem, but to distill the single most critical "Core Insight" or "Aha! Moment" required to find the solution.

* Core Task

You will be given a [Question] and its final [Answer]. 
Your sole job is to reverse-engineer the most likely solution path and identify the crucial "mental bridge"—the non-obvious insight, change in perspective, or core principle—that unlocks the problem.

* Thinking Framework

Analyze the Gap: First, understand the [Question] and look at the [Answer]. 
The core difficulty lies in the conceptual space between them. 
What makes bridging this gap non-trivial?
Reconstruct the "Hidden" Step: Mentally construct the most elegant solution path. 
In that path, what is the single most pivotal, non-obvious leap of logic or application of a principle that a student is most likely to miss?
Distill the Insight: Condense this pivotal leap into an extremely short, potent, and core-focused sentence. This sentence is the key that unlocks the door, not the map of the room.

* Constraints

Absolute Brevity: The insight must be a single sentence, ideally under 20 words.
No Spoilers: The insight must not reveal any part of the [Answer] or the specific numbers used to calculate it.
Inspirational, Not Instructional: It should inspire thought ("heuristic"), not provide a step-by-step recipe ("algorithmic").
Target the Crux: It must address the most critical linchpin that makes the entire solution possible.

* Output Format

Directly output the single, distilled "Core Insight". Do not include any other explanations, headings, or conversational text.\vspace{1em}

\textbf{User:} 

\#\#\# Question: 

[Question]

\#\#\# Answer: 

[Answer]
\end{tcolorbox}

\begin{tcolorbox}
[title=Prompt Template for Generating Ground Truth, colback=white, colframe=black!75!white, breakable]
\linespread{1.3}\selectfont
\textbf{System:} You are a helpful AI assistant. A conversation takes place between the User and the Assistant. The User asks a question, and the Assistant solves it. Please help me solve this question. Wrap only the final answer in \textit{\textbackslash \textbackslash boxed\{\}}. \vspace{1em}

\textbf{Question:} [Question] \vspace{1em}

\textbf{User:}
\end{tcolorbox}

\begin{tcolorbox}
[title=Prompt Template for Evaluation, colback=white, colframe=black!75!white, breakable]
\linespread{1.3}\selectfont
\textbf{System:} You are a helpful AI assistant. A conversation takes place between the User and the Assistant. The User asks a question, and the Assistant solves it. Please help me solve this question. Wrap only the final answer in \textit{\textbackslash \textbackslash boxed\{\}}. \vspace{1em}

\textbf{Question:} [Question] \vspace{1em}

\textbf{User:}
\end{tcolorbox}

\clearpage
\section{Further Analysis}
\label{sec:further analysis}

\subsection{UC Analysis}
\label{app:uc_analysis}
\begin{figure}[h]
  \centering
  \includegraphics[width=0.65\columnwidth]{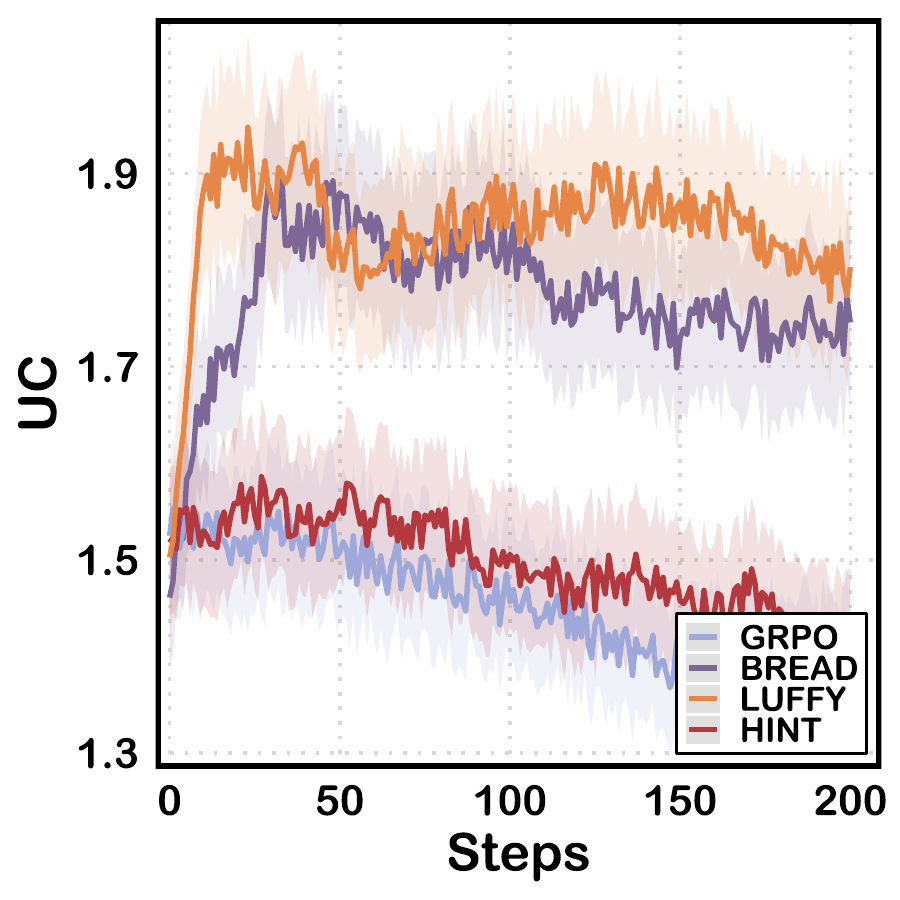} 
  \caption{Analysis of Update Consistency (UC). HINT exhibits low UC, mirroring the stability of on-policy GRPO, whereas baselines show significant variance spikes.}
  \label{fig:uc_appendix}
\end{figure}
In addition to EUR and Affinity, we analyzed Update Consistency (UC) to evaluate the variance of gradient estimates.
As shown in Figure~\ref{fig:uc_appendix}, there is a clear contrast in stability between the methods.
\textbf{HINT maintains on-policy-level stability.}
Baselines like BREAD and LUFFY quickly spike to high UC values with significant variance, reflecting unstable gradient estimates caused by large importance sampling weights.
Remarkably, the UC curve of HINT remains low and stable, almost overlapping with that of standard GRPO.
This demonstrates that despite incorporating external guidance, HINT preserves the low-variance training dynamics characteristic of on-policy learning, thereby guaranteeing convergence stability.

\subsection{Details of HINT's Entropy}
\begin{figure}[h]
  \centering
  \includegraphics[width=\linewidth]{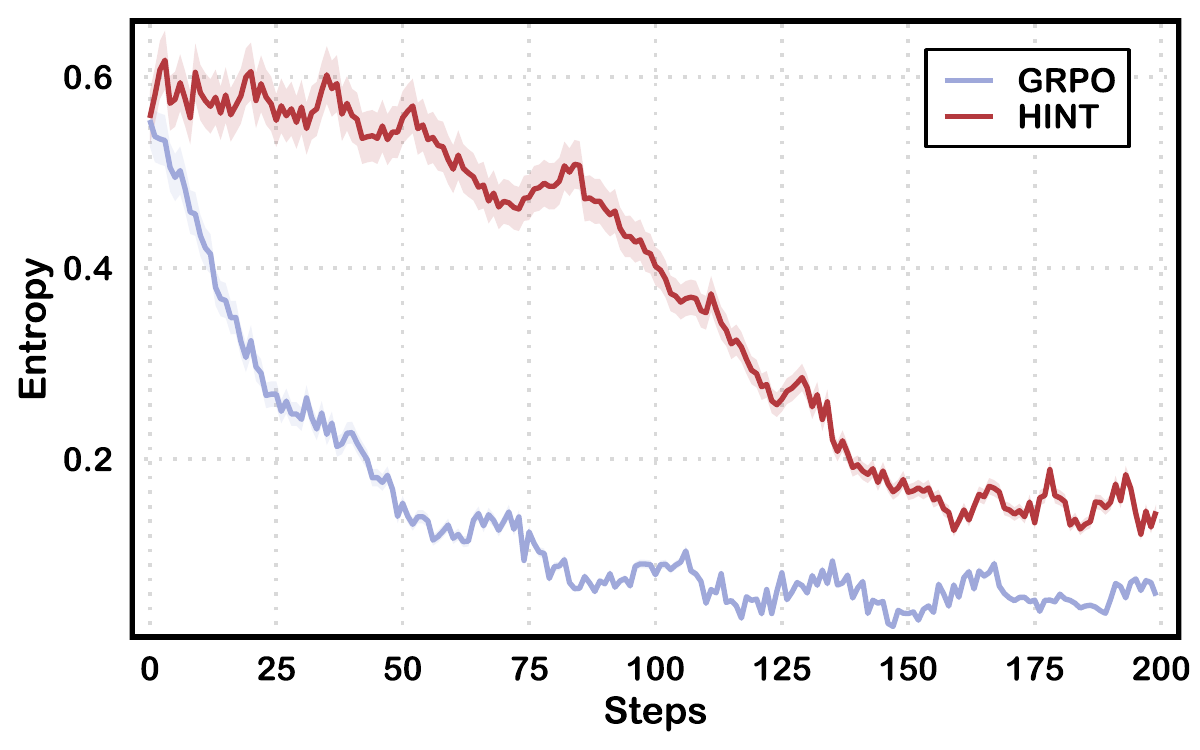}
  \caption{\textbf{HINT Prevents Entropy Collapse and Encourages Sustained Exploration.} HINT maintains a high entropy level, especially in the early stages, and stabilizes at a significantly higher value. This demonstrates that HINT's heuristic guidance fosters more continuous and diverse exploration, preventing premature policy convergence.}
  \label{fig:entropy}
\end{figure}
\textbf{HINT Encourages Sustained Exploration.}
The entropy of the generation distribution serves as a key indicator of exploration diversity. 
As illustrated in Figure~\ref{fig:entropy}, HINT avoids the rapid entropy collapse observed in GRPO during the early stages of training. 
Instead, HINT maintains a consistently high level of entropy, indicating that the model actively explores when first introduced to the hints. This period of high exploration corresponds directly to the ``EUR collapse'' phase (discussed in Section~\ref{sec:affinity_analysis}), explaining that while the model initially resists the off-policy guidance, it is nevertheless engaged in a productive and diverse search of the solution space.

During the middle stages of training, HINT's entropy does not decrease monotonically. 
It exhibits periodic increases. We attribute this to the model encountering novel types of hints and adapting its exploratory behavior to learn how to utilize them. 
Crucially, even after the policy stabilizes in the later stages, HINT maintains a significantly higher entropy level than GRPO. 
This provides strong evidence that HINT's heuristic guidance successfully fosters more continuous and diverse exploration, preventing the policy from prematurely converging to a deterministic state.

\subsection{Details of HINT's Accuracy}
\begin{figure}[h]
  \centering
  \includegraphics[width=\linewidth]{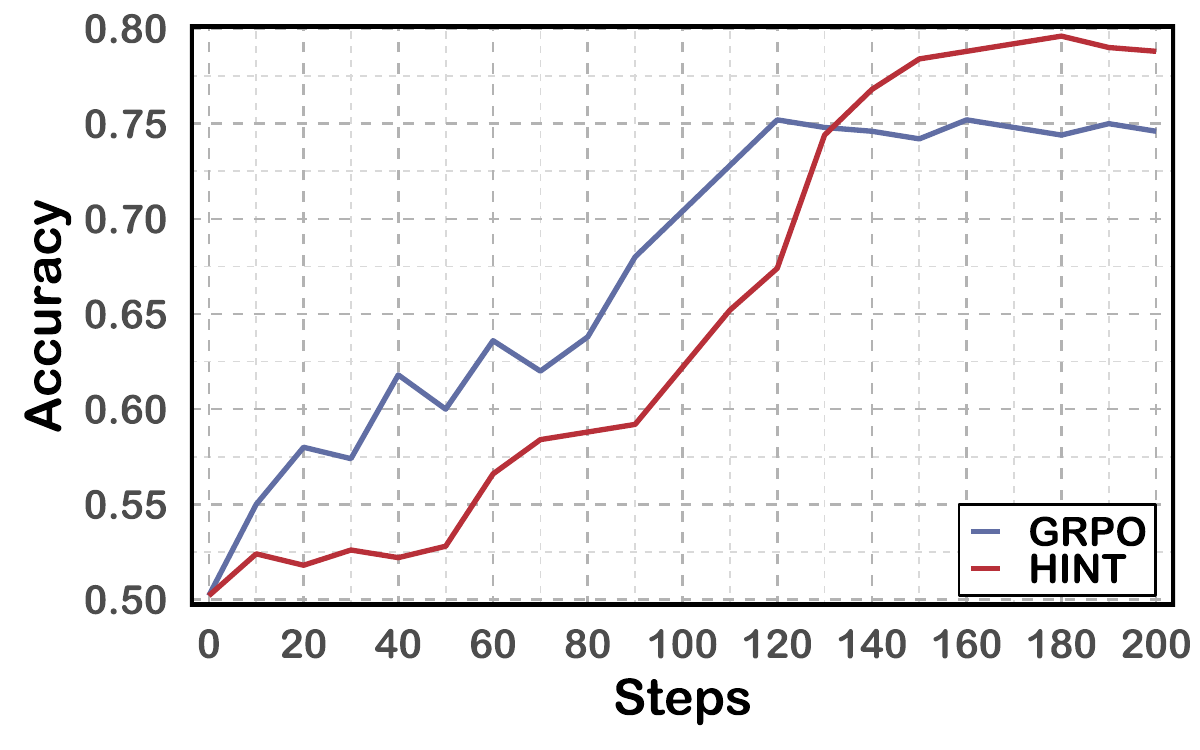}
  \caption{Accuracy of Different Methods. \textbf{HINT Achieves Higher Final Accuracy Despite Slower Initial Convergence.}}
  \label{fig:Acc}
\end{figure}

Our results reveal an interesting trade-off: while the off-policy guidance from HINT may initially slow the rate of convergence, it ultimately enables the model to achieve a higher performance ceiling. 
As shown in Figure~\ref{fig:Acc}, HINT initially exhibits a slower rate of accuracy improvement compared to GRPO. 
This initial lag is consistent with the early training stages where the model shows resistance to the heuristic hints and has not yet learned to leverage them effectively. 
However, as training progresses, the model begins to adapt and utilize the guidance. 
This leads to an accelerated learning rate after approximately 100 steps, with HINT's accuracy eventually surpassing GRPO's and reaching a higher final value. 
This dynamic suggests that the model requires an adaptation period to master the use of heuristic hints, but once learned, this skill allows it to develop stronger and more robust capabilities.

\section{LLM Usage}
Large Language Models (LLMs) were used to aid in the writing and polishing of the manuscript. Specifically, we used an LLM to assist in refining the language, improving readability, and ensuring clarity in various sections of the paper. The model helped with tasks such as sentence rephrasing, grammar checking, and enhancing the overall flow of the text.

It is important to note that the LLM was not involved in the ideation, research methodology, or experimental design. All research concepts, ideas, and analyses were developed and conducted by the authors. The contributions of the LLM were solely focused on improving the linguistic quality of the paper, with no involvement in the scientific content or data analysis.

The authors take full responsibility for the content of the manuscript, including any text generated or polished by the LLM. We have ensured that the LLM-generated text adheres to ethical guidelines and does not contribute to plagiarism or scientific misconduct.

\clearpage
\subsection{Case Study}
\begin{tcolorbox}[
    title=Case Study, 
    colback=white, 
    colframe=black!75!white, 
    width=\textwidth, 
    boxrule=1pt,  
    before upper={\linespread{1.2}\selectfont} 
]
    \textbf{Question:} Let $a$ and $b$ be positive integers such that all but $2009$ positive integers are expressible in the form $ma + nb$, where $m$ and $n$ are nonnegative integers. If $1776$ is one of the numbers that is not expressible, find $a + b$.

    \vspace{0.5em}
    \hrule 
    \vspace{0.5em}

    \textbf{Meta-Hint:} How many positive integers cannot be expressed in the form $ma+nb$ with nonnegative integers $m,n$?

    \vspace{1em}

    \begin{minipage}[t]{0.48\linewidth} 
        \raggedright 
        \textbf{Before the Hint:}
        
        By the two-coin Frobenius result, for coprime positive integers $a,b$ the number of positive integers not representable as $ma+nb$ equals $\dfrac{(a-1)(b-1)}{2}$.
        
        So 
        \[
        \begin{aligned}
        \frac{(a-1)(b-1)}{2} &= 2009 \\
        (a-1)(b-1) &= 4018
        \end{aligned}
        \]
        
        Factor $4018=41\cdot 98$. Take $a=42, b=99$.  
        Hence $a+b=141$.
        
        \vspace{1em}
        \textbf{Answer: } $\boxed{141}$ {\color{red}\XSolidBold} 
    \end{minipage}
    \hfill \vrule \hfill 
    \begin{minipage}[t]{0.48\linewidth}
        \raggedright
        \textbf{After the Hint:}
        
        By the two-coin Frobenius result, for coprime positive integers $a,b$ the number of positive integers not representable as $ma+nb$ equals $\dfrac{(a-1)(b-1)}{2}$.
        
        So 
        \[
        \begin{aligned}
        \frac{(a-1)(b-1)}{2} &= 2009 \\
        (a-1)(b-1) &= 4018
        \end{aligned}
        \]
        
        Candidates for $(a-1, b-1)$: 
        $(2,4019)$, $(3,2010)$, $(8,575)$,
        $(15,288)$, $(42,99)$, $(50,83)$.
        
        Require $\gcd(a,b)=1$, leaving:
        $(2,4019)$, $(8,575)$, $(50,83)$.
        
        Check $1776$: representable for first two, not for $(50,83)$.  
        
        Thus $a+b=133$.
        
        \vspace{1em}
        \textbf{Answer: } $\boxed{133}$ {\color{green}\CheckmarkBold}
    \end{minipage}
\end{tcolorbox}

\end{document}